%% file: HESGAforHPOGNN.tex
\documentclass[journal]{IEEEtran}
\ifCLASSINFOpdf
\else
\fi
\usepackage{graphicx}
\usepackage{booktabs}
\usepackage{float}
\usepackage{lipsum}
\usepackage{amsmath}
\usepackage{multirow}
\usepackage{subcaption}
\usepackage{listings}
\usepackage{algorithm} 
\usepackage{algorithmic}
\begin{document}
\title{A Novel Genetic Algorithm with Hierarchical Evaluation Strategy for Hyperparameter Optimisation of Graph Neural Networks}

\author{Yingfang Yuan\IEEEauthorrefmark{2}, Wenjun Wang\IEEEauthorrefmark{2}, George M. Coghill, Wei Pang\IEEEauthorrefmark{1}% <-this % stops a space
\thanks{Y.F. Yuan, W.J. Wang and W. Pang are with the School of Mathematical and Computer Sciences, Heriot-Watt University, Edinburgh, EH14 4AS, UK.}% <-this % stops a space
\thanks{G.M. Goghill is with the School of Natural and Computer Sciences, University of Aberdeen, Aberdeen, AB24 3UE, UK.}% <-this % stops a space
\thanks{Manuscript received XX XX, 2020; revised XX, XX, 20XX.}
\thanks{{\IEEEauthorrefmark{2}} Equal contribution (co-first authors)}
\thanks{{\IEEEauthorrefmark{1}} Corresponding author: Wei Pang (email: w.pang@hw.ac.uk)}
}

\maketitle

\begin{abstract}
Graph representation of structured data can facilitate the extraction of stereoscopic features, and it has demonstrated excellent ability when working with deep learning systems, the so-called Graph Neural Networks (GNNs). Choosing a promising architecture for constructing GNNs can be transferred to a hyperparameter optimisation problem, a very  challenging task due to the size of the underlying search space and high computational cost for evaluating candidate GNNs. To address this issue, this research presents a novel genetic algorithm with a hierarchical evaluation strategy (HESGA), which combines the full evaluation of GNNs with a fast evaluation approach. By using full evaluation, a GNN is represented by a set of hyperparameter values and trained on a specified dataset, and root mean square error (RMSE) will be used to measure the quality of the GNN represented by the set of hyperparameter values (for regression problems). While in the proposed fast evaluation process, the training will be interrupted at an early stage, the difference of RMSE values between the starting and interrupted epochs will be used as a fast score, which implies the potential of the GNN being considered. To coordinate both types of evaluations, the proposed hierarchical strategy uses the fast evaluation in a lower level for recommending candidates to a higher level, where the full evaluation will act as a final assessor to maintain a group of elite individuals. To validate the effectiveness of HESGA, we apply it to optimise two types of deep graph neural networks. The experimental results on three benchmark datasets demonstrate its advantages compared to Bayesian hyperparameter optimization.
\end{abstract}


% \fi
\IEEEpeerreviewmaketitle

\section{Introduction}
\label{section:1 introduction}
\IEEEPARstart{G}{raph} can be used to represent features of structured data. Deep learning equipped with graph models, the so-called graph deep learning approaches, have recently been used to predict molecular and polymer proprieties \cite{wu2018moleculenet}, and tremendous success has been achieved in comparison to the traditional approaches based on semantic SMILES strings \cite{weininger1988smiles} only. Among many types of graph deep learning systems, Graph Convolutional Neural Networks (GNNs) succeed in deep learning with promising performance and scalability \cite{long2020graph}. Generally, a GNN models a set of objects (nodes) as well as their connections (edges) in the form of topological graphs using stereoscopic features  \cite{zhou2018graph}, which is distinct from traditional vector-based machine learning systems. Thus, GNNs are good at solving graph-related problems, and it can deal with complex real-world systems in an end-to-end manner \cite{cai2018comprehensive}. Technically, GNNs can operate directly on graphs, while in molecular and polymer property prediction problems, there is a common representation transfer module which bridges the gap between the SMILES strings and graphs \cite{duvenaud2015convolutional}. Fed by the graphs, GNNs can learn to approximate the desirable properties of molecules or polymers on various user-specific scenarios or applications.

As with most of the machine learning approaches, GNNs also need a set of hyperparameters to shape their architectures, and examples of hyperparameters include the numbers of convolutional layers, filters (kernels), full connected nodes and training epochs. These hyperparameters will affect the training and learning performance, i.e., a good configuration of hyperparameters for a GNN will lead to effective training and accurate predictions, while a poor configuration will generate otherwise results. Therefore, hyperparameter optimisation (HPO) for GNN architectures is vital. Recently, Nunes et.al. \cite{nunes2020neural} compared reinforcement learning based and evolutionary algorithms based methods for optimising GNN architectures. Moreover, GraphNAS \cite{gao2020graph} employs a recurrent network trained with the policy gradient to explore network architecture. However, in the context of GNN, the HPO research is still growing \cite{nunes2020neural}. 

% There are several important research works along this line, e.g. networks can be generated by motif-based \cite{huang2017densely} or morphism-based methods \cite{wei2016network}, and  However, in the context of GNN, the HPO research is sitll growing.  \cite{nunes2020neural}. 

On the other hand, compared with traditional machine learning methods, most of deep learning models including GNNs have more sophisticated architectures and are more time-consuming to train. This means HPO for GNN is indeed a very expensive task: for each trial regarding a configuration of hyperparameters, it has to complete the full training process to evaluate the quality of this configuration. Existing HPO methods include grid search \cite{claesen2015hyperparameter}, random search \cite{schumer1968adaptive}, Gaussian and Bayesian methods \cite{snoek2012practical}, as well as evolutionary approaches \cite{di2018genetic} \cite{orive2014evolutionary} \cite {xiao2020efficient}, however most of these suffer from the expensive computational cost.

To address the expensive HPO problem, in this research we aim to develop a novel genetic algorithm (GA) with two evaluation methods: full and fast evaluations. Regarding the full evaluation, GNN will be trained on a specified dataset given a set of hyperparameter values, and root mean square error (RMSE) on the validation set will be considered as a full score of this solution. The proposed fast evaluation approach employs the difference of RMSEs between the early stage and the beginning of training as the fitness score to approximate the performance of a GNN being fully trained. A hierarchical evaluation strategy named HES is also proposed for coordinating these two evaluation methods, in which the fast evaluation operates in a lower level for recommending candidates, thereafter the full evaluation will act as a final assessor to maintain a group of elite individuals. Finally, the above procedures and operations constitute the proposed algorithm termed HESGA.  

To assess the effectiveness of the proposed HESGA, we carried out experiments on three public molecule datasets: EOSL \cite{delaney2004esol}, FreeSolv \cite{mobley2014freesolv}, and Lipophilicity \cite{wenlock_experimental_2015}, which involve the predictions of three properties: molecular solubility, hydration free energy, and lipophilicity, respectively. Each dataset has all the molecules represented by SMILES strings,  which can be used to construct molecular graphs. These constructed graphs were then used as input to GNNs for the predictive tasks. In this research, we apply HESGA to optimise the hyperparameters of two types of Graph Neural Networks: Graph Convolution (GC) \cite{duvenaud2015convolutional} and Message Passing Neural Network (MPNN) \cite{vinyals2015order}, to improve their learning performance in terms of RMSE. The promising results compared with benchmarks \cite{wu2018moleculenet} show that HESGA has advantages in both achieving good/better solutions compared to the Bayesian based HPO and require less computational cost than the original genetic algorithm.

The main contributions of this research are as follows:
\begin{enumerate}
    \item We proposed a fast approach for evaluating GNNs by using the difference of RMSEs between the early training stage and the very beginning of the training. 
    \item We proposed a novel hierarchical evaluation strategy used together with GA (HESGA) for hyperparameter optimisation. 
    \item We conducted systematic experiments on three benchmark datasets (ESOL, FreeSolv, and Lipophilicity) to assess the performance of HESGA on GC and MPNN models as opposed to the Bayesian method.
\end{enumerate}

The rest of this paper is organized as follows. Section \ref{section: 2 background} introduces relevant work and methods for HPO. In Section \ref{section: 3 HESGA}, the details of HESGA is presented. The experiments are reported and the results are analysed in Section \ref{section: 4 Experiments}. Then regarding more interesting topics we have a further discussion in Section \ref{section: 5 further discussion}. Finally, Section \ref{section: 6 conclusion} concludes the paper and explores some directions in future work.

\input{Section2Background}
\input{Section3HESGA}

\input{Section4Experiments}
\input{Section5FurtherDiscussion}
\input{Section6Conclusion}
\input{Section7Acknowledge}
\ifCLASSOPTIONcaptionsoff
  \newpage
\fi



\bibliographystyle{IEEEtran}
\bibliography{mybib}
\end{document}

%% file: Section2Background.tex
\section{Background and Relevant Methods}
\label{section: 2 background}
\subsection{Grid Search and Random Search}
Grid search and random search are two most commonly used approaches for HPO. Grid-based method initializes the hyperparameter space using grid layout and tests each point (representing a configuration of hyperparameters) in the grid. As grid search is performed in an exhaustive manner to evaluate all grid points, its cost is determined by the resolution of the pre-specified grid layout. In contrast, random search is supported by a series of defined probabilistic distributions (e.g. uniform distribution), which suggest a number of points in trials. It is noted that random search is in general more practical and efficient than grid search for HPO of neural networks given the same computational budget \cite{bergstra2012random}.

\subsection{Bayesian and Gaussian Approaches}
Bayesian optimization can be used to suggest the probabilistic distributions mentioned in random search. It is assumed that the performance of the learning model is correlated to its hyperparameters. Thus, we give a higher probability to the set of hyperparameter values with better performance \cite{snoek2012practical}, which means that it will be allocated with more chances to be sampled further. After sufficient iterations of calculations, a probability distribution function similar to the maximum likelihood function can be learned by Bayesian approaches \cite{wistuba2018scalable}, random forest \cite{eggensperger2013towards} and other surrogate models. As a result, the computational cost for model validation will be saved in this way. Gaussian process is suitable for approximating the distribution of evaluation results because of their flexibility and traceability \cite{snoek2015scalable}. The combination of Bayesian optimisation with Gaussian process outperforms human expert-level optimization in many problems \cite{snoek2012practical}. For example, in \cite{klein2017fast}, FABOLAS is poposed to accelerate Bayesian optimisation of hyperparameters on large datasets, and this method benefits from sub-sampling. Another successful case is a method which combines Bayesian optimisation and Hyperband, and it possesses the features of simplicity, efficiency, robustness and flexibility \cite{falkner2018bohb}.

\subsection{Evolutionary Computation }
Recent years, evolutionary algorithms (EAs) have demonstrated advantages in solving large-scale, highly non-linear and expensive optimisation problems \cite{deb2001multi} \cite{coello2007evolutionary}. HPO for GNNs is usually expensive \cite{8909403} in evaluating each of the feasible architectures. Thus, using EAs to solve HPO problems have been explored due to their excellent search ability \cite{young2015optimizing}.

For using EA, the representation of solutions (encoding) is a key issue, for which direct acyclic graphs \cite{suganuma2017genetic} and binary representation \cite{Xie_2017_ICCV} have demonstrated their advantages. Given good representations of hyperparameters, EA will generate a population of individuals as potential solutions, each of which will be evaluated by a fitness function. In most cases the fitness function is the objective function, which in our context means first fully training a GNN with the specified hyperparameters and then evaluating its learning performance ( (in terms of RMSE)) as the fitness value. Thereafter, these individuals are selected by a selection method (e.g., the roulette) based on their fitness values to be the parents. Through evolutionary iterations, those GNNs with higher fitness values are more likely to be maintained in the population, and those fitter solutions will have more chance to produce offspring. In the end, the best individual will be selected as the final GNN model.

There are two main issues in evolutionary computation: convergence of the algorithm and diversity of population. To make the evolutionary search converge faster, researchers have proposed many methods, including modification of evolutionary operators \cite{zhu2016novel}, using elite archive \cite{zhu2017external}, ensembles \cite{wang2018effective} to increase the chance of selecting better parents, and niching methods \cite{lin2016adaptive} for local exploitation \cite{li2015pareto}. In terms of population diversity, some approaches have designed for escaping local optima  and improving the performance of exploration \cite{yang2016novel}. Regarding the above two issues, in this research we will propose a novel GA with an elite archive for increasing convergence and a mating selection strategy which allows one parent to be selected from the whole population for increasing diversity.

%% file: Section3HESGA.tex
\section{Genetic Algorithm with Hierarchical Evaluation Strategy}
\label{section: 3 HESGA}
In many real-world applications GNN suffers from expensive computational cost, so HPO for GNN is a challenging task, particularly in those cases with huge hyperparameter search space. Moreover, GA maintains a population of individuals (as solutions) during the search, which means in one generation the computational cost may involve evaluating all GNN models in the population. To address this issue, a surrogate model with lower evaluation cost \cite{chugh2020surrogate} or a faster evaluation method \cite{frachon2019immunecs} can be considered. However, there is no guarantee that the fitness values generated by such methods would reliably approximate those obtained from the original evaluation function, and therefore the HPO results based on such methods may be poor. A good idea is to combine both the original and fast evaluation strategies together in GA, in which case we can achieve a tradeoff between performance and computational cost.

In the rest of this section we will first introduce the motivation of HESGA, and then we will present the following two detailed processes: (1) fast evaluation by using difference of RMSEs and (2) the hierarchical evaluation strategy. Next, the full HESGA is presented with a scalable module for fast evaluation. At last, the settings for HESGA are presented.

\subsection{ Solution Encoding }\label{section:3.1}
Take an example by four of the hyperparameters mentioned in the benchmark problems: batch size ($s_{b}$), the number of filters in convolution layer ($n_{f}$), learning rate ($r_{l}$) and the number of fully connected nodes ($n_{n}$). A binary encoding for these four hyperparameters is shown in Table \ref{tab:Encodes for Hyperparameter and Solution}.

\begin{table*}[!t]
\renewcommand{\arraystretch}{2}
\caption{Encoding for Hyperparameter and Solution}
\centering
\begin{tabular}{|l|l|l|l|l|}
\hline
 & $s_{b}$ & $n_{f}$ & $r_{l}$ & $n_{n}$\\ \hline
\textbf{Binary encoding} & {[}0 0 0{]}$\sim${[}1 1   1{]} & {[}0 0 0{]}$\sim${[}1 1   1{]} & {[}0 0 0 0{]}$\sim${[}1   1 1 1{]} & {[}0 0 0{]}$\sim${[}1 1   1{]} \\ \hline
\textbf{Range of hyperparameters} & 1$\sim$8 & 1$\sim$8 & 1$\sim$16 & 1$\sim$8 \\ \hline
\textbf{Resolution (step increment)} & 32 & 32 & 0.0001 & 64 \\ \hline
\textbf{Full integer ranges} & 32$\sim$256 & 32$\sim$256 & 0.0001$\sim$0.0016 & 64$\sim$512 \\ \hline
\end{tabular}
\label{tab:Encodes for Hyperparameter and Solution}
\end{table*}

\begin{figure}[!t]
    \centering
    \includegraphics[width=3.5in]{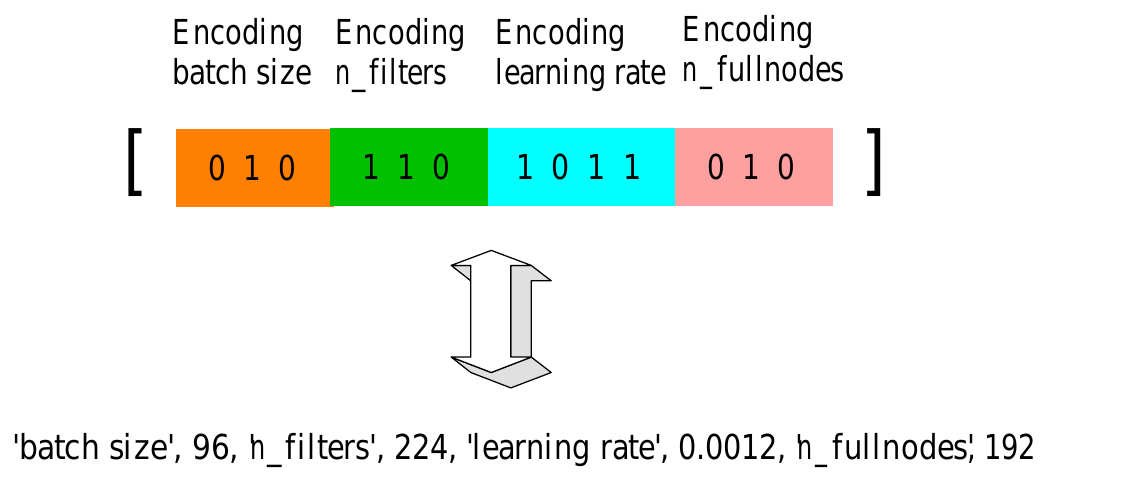}
    \caption{An Example Encode of Solution}
    \label{fig:fig1}
\end{figure}

In Table \ref{tab:Encodes for Hyperparameter and Solution}, three 3-bit binary strings are used to represent the parameters: $s_{b}$, $n_{f}$, and $n_{n}$, together with the resolutions of 32, 32, and 64, respectively according to the benchmark problems. A 4-bit binary string is used to represent the learning rate ($r_{l}$) with a resolution (step increment) of  0.001 accordingly. Thus, we have the feasible ranges for batch size as $[32\sim256]$, the number of filters as $[32\sim256]$, learning rate as $[0.0001\sim0.0016]$ and the number of fully connected nodes as $[64\sim512]$. It is noted that because the binary string “$000$” corresponds to the decimal integer $0$, but $0$ is not expected by all of the hyperparameters. So we transfer the mapping from binary  to decimal integer by adding the value of 1 upon the decimal integer, e.g. $000$ will be mapped as decimal integer $1$, $001$ is mapped to $2$, and $111$ will correspond to integer $8$. According to Table \ref{tab:Encodes for Hyperparameter and Solution}, an example of encoding a solution is shown in Fig. \ref{fig:fig1}. 

With the encoding strategy specified, EA will be able to perform effective search for the optimal individual in the hyperparameter space. In deed, the performance of EA will be affected by many factors, such as population size, maximum number of generations, operators for producing offspring, and population maintenance strategy, etc. However, we believe that with common parameters, EA will reach the optimal solution with less evaluation times than grid search and random search \cite{8297018}.

\subsection{Full Evaluation and Fast Evaluation}
 Regarding full evaluation, a GNN is first represented by a set of hyperparameter values and then trained on a specified dataset. At the end of training, the trained GNN will be validated on another specified dataset, and the RMSE of the validation will be used to measure the quality of the set of hyperparameter values as full evaluation.

There are already several approaches on developing fast evaluations, such as partial training \cite{frachon2019immunecs} \cite{zoph2018learning}, and incomplete training \cite{zela2018towards} \cite{real2019regularized}. Partial training with a sub-dataset is good at tackling big datasets and complicated models. However, when the dataset is not very big, e.g. the dataset FreeSolv \cite{mobley2014freesolv}  with only $642$ data points, partial training seems not appropriate due to the lack of data points for training. While the incomplete training with early stop policy might be helpful for processing such datasets like FreeSolv. Based on these ideas, a fast evaluation method by using the difference of RMSEs of validation between the early stage and the very beginning of training is introduced. In the below Equation \ref{eq1}, \textit{F(t)} stands for the fitness value at epoch \textit{t} during GNN training. 

\begin{equation*}
    \label{eq1}
    \tag{1}
    \Delta F(1, t)=F(1)-F(t).
\end{equation*}

In the above, $\Delta F(1, t)$ is defined as the difference of fitness between the $1^{st}$ epoch and $t^{th}$epoch. In our experiments, RMSE was used as the fitness evaluation metric, so $\Delta F(1,t)$ can approximate the rate of decrease in RMSE. As a heuristic, those individuals in the population with bigger $\Delta F(1,t)$ values are more promising to achieve smaller RMSE at the end of their training. We note that this may not be always the case, but we use this as an approximate value for the final fitness value in order to reduce the cost for evaluating GNNs. We also note that the number of $t$ epochs will be far less than the number of epochs needed in training, so $\Delta F(1,t)$ can also be called difference fitness in the early training stage. 

By using this difference fitness, we can offer a fast evaluation to all individuals in the population, according to their performance in the early training stage. However, there is a key issue that needs to be addressed: how to choose the argument $t$. Since some training algorithms would terminate the training by a fixed maximum number of epochs, while the others might have a more adaptive criterion for termination, we cannot set a fixed argument $t$ for the fast evaluation. So $10 \%^{\sim} 20 \%$ of the maximum number of epochs is proposed, which means the fast evaluation will only consume approx. $10 \%^{\sim} 20 \%$ of the computational cost compared to the full evaluation. 

\subsection{Hierarchical Evaluation Strategy}
\begin{figure}[!t]
    \centering
    \includegraphics[width=3.6in]{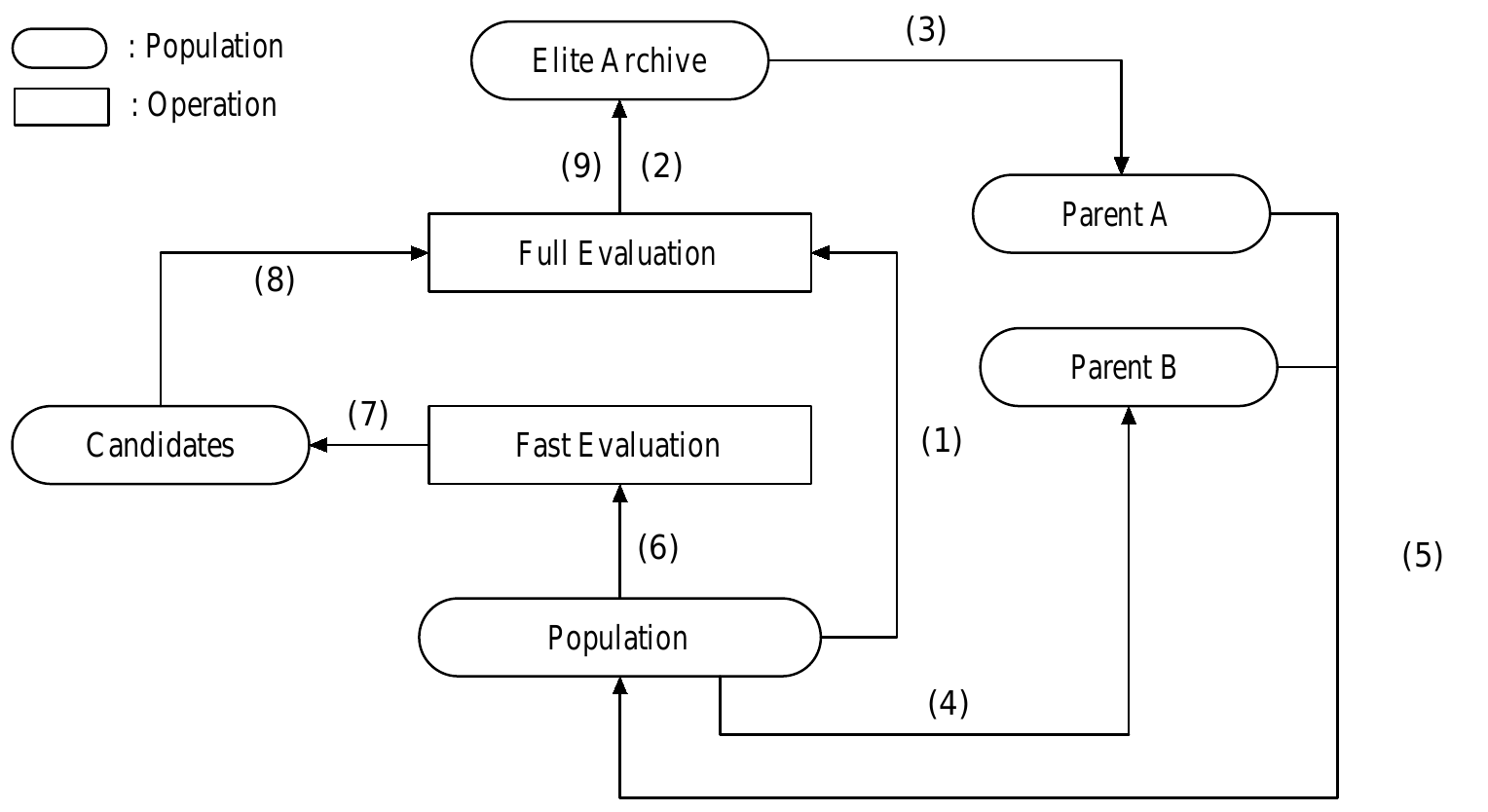}
    \caption{Hierarchical Evaluation Strategy}
    \label{fig:fig2}
\end{figure}

The fast evaluation will only suggest the individuals which have high probability of achieving better results after full training, but it still cannot guarantee that this is always the case. Thus, a hierarchical structure including both fast and full evaluations is designed as shown in Fig. \ref{fig:fig2}.

In Fig. \ref{fig:fig2}, after population initialization, all the individuals are assessed by the full evaluation method in Step (1), and in Step (2), those with higher fitness values are selected and sent to the elite archive accordingly. In Steps (3) and (4), parents A and B are individually selected by the roulette method from the elite archive and the whole population. A new population are generated in Step (5) to replace the old one. In Step (6), all the individuals in the new population will not take the full evaluation, alternatively they are assessed by the fast evaluation method, and a small number of candidates with better fitness values will be selected in Step (7). Further, these candidates are assessed by the full evaluation method in Step (8), and then in Step (9), they will update the elite archive depending on if they are better than some of the individuals in the elite archive. Next, Steps (3) and (4) will be repeated to generate new offspring, and the whole process will be run iteratively until the termination criteria are met. 

\subsection{Full HESGA and Parameter Settings}
The pseudo code of HESGA and the parameter settings are shown in Algorithm \ref{alg1}.

\begin{figure}[!t]
\begin{algorithm}[H]
\caption{HESGA}
\label{alg1}
\begin{algorithmic}[1]
    \STATE \textbf{Initialization} with solution and population, $n_{pop}$, ${d_{indi}}$\;
    
    \STATE \textit{gen = 0}, $maxgen$, $r_{e}=0.1$, $r_{c}=0.1$, $p_{c}=0.8$, $p_{m}=0.2$, $ev_{fast}=0$, $ev_{full}=0$ \;
    
    \STATE population evaluated by full evaluation, update the elite archive, $ev_{full} += n_{pop}$,  \; 
    
    % \WHILE{gen $<$ max_gen}
    %     \STATE select parent A and B from elite archive and the whole population, respectively to generate n\_pop new offspring \; 
    %     \STATE lower evaluation on whole population, then select better ones to candidate group, ev\_fast += n\_pop, \;
    %     \STATE higher evaluation on candidate group, to update elite archive, ev\_full +=n\_pop * r\_e, \;
    %     \STATE save the best individual of elite archive, gen ++ \;
    % \ENDWHILE
    \WHILE {$gen < maxgen$}    
        \STATE select Parents A and B from the elite archive and the whole population, respectively, to generate $n_{pop}$ new offspring \; 
        \STATE fast evaluation on the new population, then select $n_{pop} \times r_{c}$ better individuals to enter the candidate group, $\mathit{ev_{fast} += n_{pop},}$ \;
        \STATE full evaluation on the candidate group, and update the elite archive, $\mathit{ev_{full} += n_{pop} \times r_{c},}$ \;
        \STATE save the best individual of elite archive, $gen ++$ \;
     
\ENDWHILE
\STATE \textbf{Output} the final GNN model decoded from the best individual in the elite archive
\end{algorithmic}
\end{algorithm}
\end{figure}

In Algorithm \ref{alg1}, $n_{pop}$ is the size of population, $d_{indi}$ is the dimension of solution which depends on the resolution as mentioned in Section \ref{section:3.1}, ${gen}$ is the counter for generations, ${maxgen}$ is the maximum number of generations allowed in one execution, $r{_e}$ and $r_{c}$ are the proportions for elite archive and candidates group, $p_{c}$ and $p_{m}$ are the probabilities for crossover and mutation, $ev_{fast}$ and $ev_{full}$ is the counter for the times of fast and full evaluations. In \textbf{Line 3}, the initial population will be evaluated by the full evaluation method to select elites, which will be sent to the elite archive. From \textbf{Line 4} to \textbf{Line 9}, the loop is executed until the termination conditions are met. In \textbf{Line 10}, the final GNN model decoded from the best individual in the elite archive will be the output.

In each loop (\textbf{Lines $\mathbf{4}\sim \mathbf{9}$}), HESGA will first assess the new offspring by fast evaluation, then the better candidates selected via fast evaluation will undergo full evaluation process as in Fig. \ref{fig:fig2}. The elite archive is then updated by the better candidates. This hierarchical evaluation strategy offers a pre-selection mechanism by the fast evaluation method proposed and could save around $80 \%^{\sim} 90 \%$ computational cost. On the other hand, the full evaluation approach acts as a final assessor, which ensures that the population moves to the right direction towards the objective function all the time. 

% During each loop, HESGA will first assess the new offspring by using lower evaluation (i.e., fast evaluation in Figure 2), for selecting candidates to be higher evaluated (i.e., by full evaluation method in Figure 2) to update the elite archive by excellent candidates. This hierarchical evaluation strategy offers a pre-selection mechanism by the lower (fast) evaluation method proposed and could save computational cost to the extent of around $80 \%^{\sim} 90 \%$ On the other hand, the higher evaluation approach act as a final examiner who will ensure that the population move to the right objective all the time. It should be noted that this idea is very common in reality, e.g., in staff recruitment a shortlist will be generated by a rougher but faster criterion firstly, and then judged by a more rigorous but time-consuming criterion. 

\subsection{Evolutionary Operators and Other Settings}

\begin{figure}[!t]
    \centering
    \includegraphics[width=3.5in]{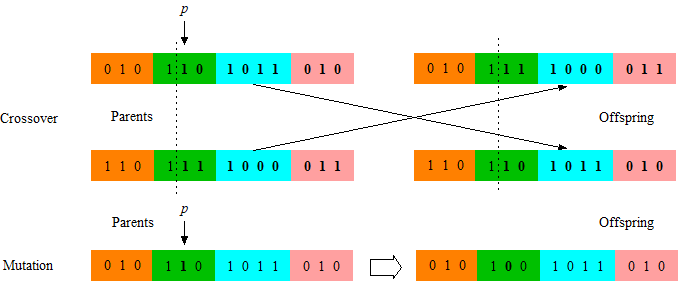}
    \caption{Binary Crossover and Mutation}
    \label{fig:fig3}
\end{figure}

% The binary crossover \cite{deb1995simulated}  \cite{lim2017crossover} and mutation \cite{mitchell1996introduction} operators are used as their original references \cite{whitley1994genetic}, their basic mechanisms are demonstrated by the follow figure.
We use the classical binary crossover and mutation operators as in \cite{deb1995simulated}  \cite{lim2017crossover} \cite{mitchell1996introduction} \cite{whitley1994genetic}, and their mechanisms are demonstrated by the example shown in Fig. \ref{fig:fig3}. In Fig. \ref{fig:fig3}, the position parameter $p$ in both crossover and mutation is a randomly generated integer in the range of $(1, \textit{len})$, where $len$ is the solution length (i.e. the number of bits in the binary string). 

The maximum generation and population size are set according to the specific problems that HESGA aims to solve. As for the population maintenance, the elite archive is maintained by the fitness sorting method, while the population does not need a maintenance policy. In elite archive update, when a better candidate can successfully update the elite archive, the worst one in the elite archives will be discarded.

%% file: Section4Experiments.tex
\section{Experiments}
\label{section: 4 Experiments}
In this section, the performance of HESGA will be experimentally investigated on several datasets mentioned in Section \ref{section:1 introduction}, and we use two types of deep graph neural architectures, Graph Convolution (GC) \cite{duvenaud2015convolutional} and Message Passing Neural Network (MPNN) \cite{vinyals2015order} to assess the performance of HESGA. Section \ref{section:4.1} shows the advantage and disadvantage of the traditional GA for HPO 
compared with the default parameter settings. Section \ref{section:4.2} presents the results obtained from optimising the GC model with the proposed HESGA compared to the Gaussian HPO method on three datasets, i.e. ESOL \cite{delaney2004esol}, FreeSolv \cite{mobley2014freesolv} and Lipophilicity \cite{wenlock_experimental_2015}. Section \ref{section: 4.3} reports the performance of HESGA on MPNN model. All experiments are performed on a PC with Inter (R) Core i5-8300 CPU, 8GB Memory, and GeForce GTX 1050 GPU.

\subsection{Advantage and Disadvantage of the Traditional GA}\label{section:4.1}
\begin{figure}[!t]
    \centering
    \includegraphics[width=2.5in]{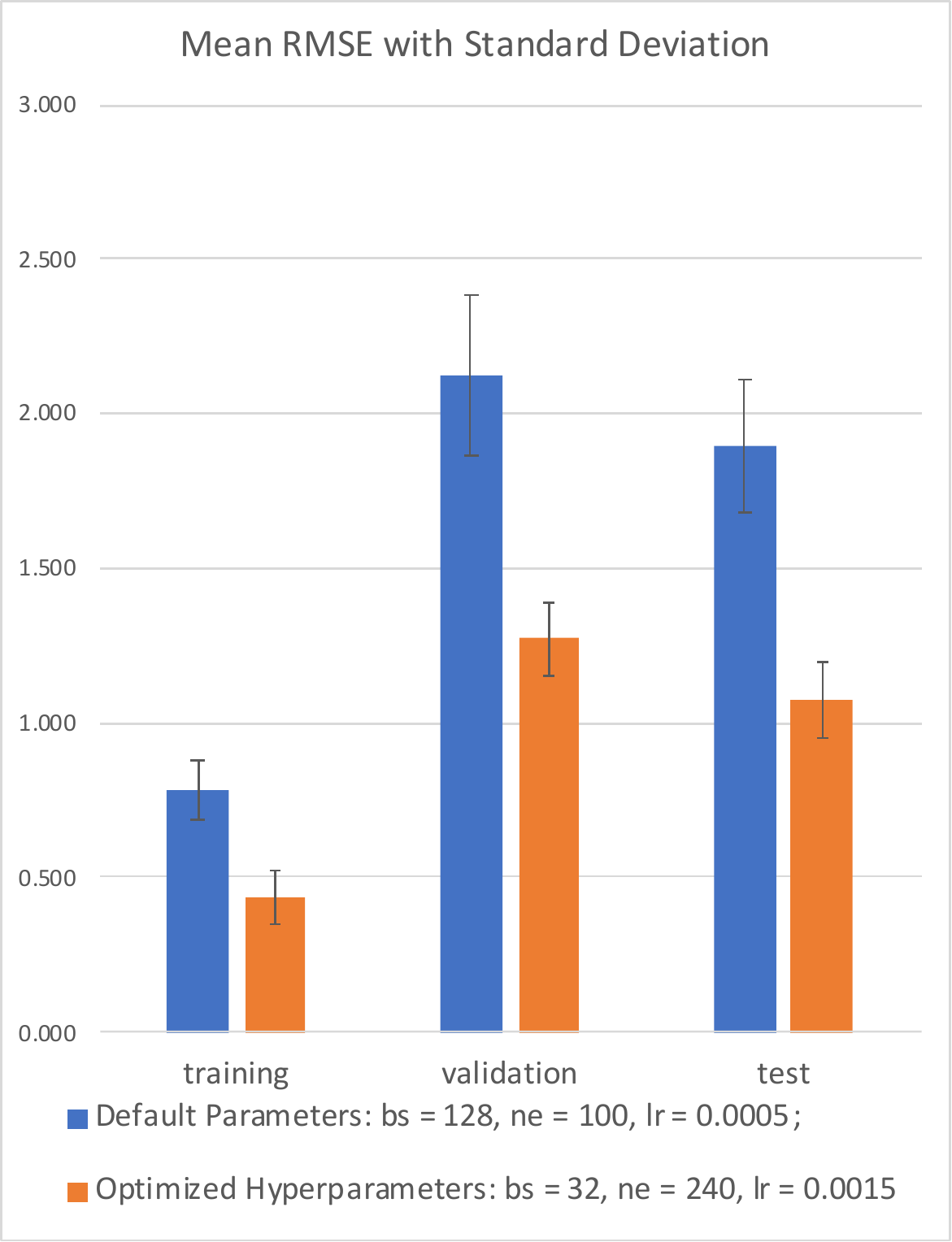}
    \caption{A comparison between GC with optimised hyperparameters and GC with default hyperparameters}
    \label{fig:fig4}
\end{figure}

For a case study on GA to optimize hyperparameters, we use a traditional GA to optimise the GC model and run it on the FreeSolv dataset. In this experiment, three parameters: batch size ($s_{b}$), the number of execution epochs ($n_{e}$), and learning rate ($r_{l}$) are optimized by GA. The hyperparameter optimized by GA are $s_{b} = 32$, $n_{e} = 240$ and $r_{l} = 0.0015$; on the other hand, the default hyperparameters pre-set in GC are $s_{b} = 128$, $n_{e} = 100$ and $r_{l} = 0.0005$. These two configurations of parameters are used to run GC for 30 times independently. The average RMSEs of training, validation and test, as well as their standard deviations are plotted in Fig. \ref{fig:fig4}. More details about the distribution of RMSE of validation will be presented in Section \ref{section:5.1}.

We carried out $t$-test on the RMSE results obtained from GC with optimized hyperparameters and GC with default hyperparameters, and it shows that these two groups of RMSEs do not have the same mean value at a significant level of 5\%, regarding training, validation and test, respectively. Thus, it is significant that using GA hyperparameter optimisation approach will improve the learning performance of GC with respect to the RMSE.

The disadvantage of the traditional GA for HPO is its intolerable computational cost, especially to those highly expensive problems. So, as mentioned in Section \ref{section: 3 HESGA}, we present HSEGA contains a fast evaluation strategy for candidate selection. 

\subsection{Experimental Results of HESGA Added on GC Model}\label{section:4.2}
To further investigating the performance of our proposed HESGA, in this section we will apply HESGA to optimise GC model, then this combination is tested on the ESOL, FreeSolv and Lipophilicity datasets. We record RMSE values for comparison with GC with Bayesian hyperparameter optimization (BHO). Each of the experiments was executed by 30 independent trials to obtain statistical results. In Tables \ref{tab:tab2 Results on ESOL Data Set (1)}$\sim$ \ref{tab:tab4 Results on Lipophilicity Data Set}, $n_{f}, n_{n}, s_{b}, n_{e}, l_{r}$ stand for the number of filters, the number of fully connected nodes, batch size, the number of maximum epoch, and learning rate, respectively. Symbol M\_ and Std\_ denote the mean and standard deviation of the corresponding RMSE, respectively. $h$ is the indicator of $t$-test, $h=1$ indicates that the hypothesis of two population groups have equal mean is rejected with a default significance level of $5\%$, in the case of which means the two group of samples have significantly different mean values. In details, M\_RMSE obtained by GC+HESGA will minus the one obtained by GC+BHO, so a negative $t$ value indicates the former is better, while a positive $t$ value indicates the former is worse.

\begin{table*}[!t]
\centering
\caption{The Results on ESOL Dataset}
\label{tab:tab2 Results on ESOL Data Set (1)}
\resizebox{\textwidth}{!}{%
\begin{tabular}{|l|l|l|l|l|l|l|l|}
\hline
\textbf{ESOL} & \textbf{Hyperparameters} & \multicolumn{2}{l|}{\textbf{Training   Results}} & \multicolumn{2}{l|}{\textbf{Validation   Results}} & \multicolumn{2}{l|}{\textbf{Test Results}} \\ \hline
\multirow{4}{*}{\textbf{GC + BHO}} & $n_{f}$ = 128 & \multirow{2}{*}{M\_RMSE} & \multirow{2}{*}{0.43} & \multirow{2}{*}{M\_RMSE} & \multirow{2}{*}{1.05} & \multirow{2}{*}{M\_RMSE} & \multirow{2}{*}{0.97} \\ \cline{2-2}
 & $n_{n}$ = 256 &  &  &  &  &  &  \\ \cline{2-8} 
 & $s_{b}$ = 128 & \multirow{2}{*}{Std\_RMSE} & \multirow{2}{*}{0.20} & \multirow{2}{*}{Std\_RMSE} & \multirow{2}{*}{0.15} & \multirow{2}{*}{Std\_RMSE} & \multirow{2}{*}{0.01} \\ \cline{2-2}
 & $r_{l}$ = 0.0005 &  &  &  &  &  &  \\ \hline
\multirow{4}{*}{\textbf{GC + HSEGA}} & $n_{f}$ = 192 & \multirow{2}{*}{M\_RMSE} & \multirow{2}{*}{0.34} & \multirow{2}{*}{M\_RMSE} & \multirow{2}{*}{0.89} & \multirow{2}{*}{M\_RMSE} & \multirow{2}{*}{0.89} \\ \cline{2-2}
 & $n_{n}$ = 448 &  &  &  &  &  &  \\ \cline{2-8} 
 & $s_{b}$ = 32 & \multirow{2}{*}{Std\_RMSE} & \multirow{2}{*}{0.07} & \multirow{2}{*}{Std\_RMSE} & \multirow{2}{*}{0.04} & \multirow{2}{*}{Std\_RMSE} & \multirow{2}{*}{0.04} \\ \cline{2-2}
 & $r_{l}$ = 0.0009 &  &  &  &  &  &  \\ \hline
\multicolumn{2}{|l|}{\begin{tabular}[c]{@{}l@{}}\textbf{T-test on results} \\
(with significance level of $\alpha$ = 5\%)\end{tabular}} & \multicolumn{2}{l|}{\begin{tabular}[c]{@{}l@{}}$t$ = -2.436, $h$   = 1 \\ to reject the \\ equal mean hypothesis\end{tabular}} & \multicolumn{2}{l|}{\begin{tabular}[c]{@{}l@{}}$t$ = -5.624, $h$ = 1 \\ to reject the \\ equal mean hypothesis\end{tabular}} & \multicolumn{2}{l|}{\begin{tabular}[c]{@{}l@{}}$t$ = -9.708, $h$ = 1 \\ to reject the \\ equal mean hypothesis\end{tabular}} \\ \hline
\end{tabular}%
}
\end{table*}

Table \ref{tab:tab2 Results on ESOL Data Set (1)}  shows very good performance of HESGA on ESOL dataset compared to the BHO approach, in which our results of average RMSE (M\_RMSE) are all significant less than those of BHO. Moreover, the hyperparameters obtained by HESGA had more stable RMSE values during 30 independent trials (i.e. less standard deviation) in both the training and validation dataset. 

\begin{table*}[!t]
\centering
\caption{The Results on FreeSolv Dataset (1)}
\label{tab:tab3 Results on FreeSolv Data Set}
\resizebox{\textwidth}{!}{%
\begin{tabular}{|l|l|l|l|l|l|l|l|l}
\cline{1-8}
\textbf{FreeSolv} & \textbf{Hyperparameters} & \multicolumn{2}{l|}{\textbf{Training   Results}} & \multicolumn{2}{l|}{\textbf{Validation   Results}} & \multicolumn{2}{l|}{\textbf{Test Results}} &  \\ \cline{1-8}
\multirow{4}{*}{\textbf{GC + BHO}} & $n_{f}$ = 128 & \multirow{2}{*}{M\_RMSE} & \multirow{2}{*}{0.31} & \multirow{2}{*}{M\_RMSE} & \multirow{2}{*}{1.35} & \multirow{2}{*}{M\_RMSE} & \multirow{2}{*}{1.40} &  \\ \cline{2-2}
 & $n_{n}$ = 256 &  &  &  &  &  &  &  \\ \cline{2-8}
 & $s_{b}$ = 128 & \multirow{2}{*}{Std\_RMSE} & \multirow{2}{*}{0.09} & \multirow{2}{*}{Std\_RMSE} & \multirow{2}{*}{0.15} & \multirow{2}{*}{Std\_RMSE} & \multirow{2}{*}{0.16} &  \\ \cline{2-2}
 & $r_{l}$ = 0.0005 &  &  &  &  &  &  &  \\ \cline{1-8}
\multirow{4}{*}{\textbf{GC + HSEGA}} & $n_{f}$ = 192 & \multirow{2}{*}{M\_RMSE} & \multirow{2}{*}{0.63} & \multirow{2}{*}{M\_RMSE} & \multirow{2}{*}{1.29} & \multirow{2}{*}{M\_RMSE} & \multirow{2}{*}{1.21} &  \\ \cline{2-2}
 & $n_{n}$ = 512 &  &  &  &  &  &  &  \\ \cline{2-8}
 & $s_{b}$ = 32 & \multirow{2}{*}{Std\_RMSE} & \multirow{2}{*}{0.12} & \multirow{2}{*}{Std\_RMSE} & \multirow{2}{*}{0.13} & \multirow{2}{*}{Std\_RMSE} & \multirow{2}{*}{0.12} &  \\ \cline{2-2}
 & $r_{l}$ = 0.0012 &  &  &  &  &  &  &  \\ \cline{1-8}
\multicolumn{2}{|l|}{\begin{tabular}[c]{@{}l@{}}\textbf{T-test on results} \\
with significance level of $\alpha$ = 5\%\end{tabular}} & \multicolumn{2}{l|}{\begin{tabular}[c]{@{}l@{}}$t$ = 12.031, $h$   = 1 \\ to reject the \\ equal mean hypothesis\end{tabular}} & \multicolumn{2}{l|}{\begin{tabular}[c]{@{}l@{}}$t$ = -2.117, $h$   = 1 \\ to reject the \\ equal mean hypothesis\end{tabular}} & \multicolumn{2}{l|}{\begin{tabular}[c]{@{}l@{}}$t$ = -5.184, $h$   = 1\\  to reject the \\ equal mean hypothesis\end{tabular}} &  \\ \cline{1-8}
\end{tabular}%
}
\end{table*}

Regarding the FreeSolv dataset, as the results shown in Table \ref{tab:tab3 Results on FreeSolv Data Set}, in the training dataset our M\_RMSE is worse than the one obtained from GC+BHO, however, in the results on validation and test datasets, our method are slightly better than GC+BHO. It should be noted that in terms of validation, these two results are very similar, as the reject $t$-value for \textit{d.f.} (degree of freedom) at 30 and at 60 are 2.042 and 2.0, as the $t$ value we got is -2.117, which is nearly on the boundary of acceptance. It is also noted that, as we do not know the exact size of RMSE sample group in the reference paper \cite{wu2018moleculenet}, we suppose it was in the range of 1$\sim$30, thus \textit{d.f.} at 30 and 60 are both considered in this work.

\begin{table*}[!t]
\centering
\caption{The Results on Lipophilicity Dataset}
\label{tab:tab4 Results on Lipophilicity Data Set}
\resizebox{\textwidth}{!}{%
\begin{tabular}{|l|l|l|l|l|l|l|l|l}
\cline{1-8}
\textbf{Lipophilicity} & \textbf{Hyperrarameters} & \multicolumn{2}{l|}{\textbf{Training   Results}} & \multicolumn{2}{l|}{\textbf{Validation   Results}} & \multicolumn{2}{l|}{\textbf{Test Results}} &  \\ \cline{1-8}
\multirow{4}{*}{\textbf{GC + BHO}} & $n_{f}$ =128 & \multirow{2}{*}{M\_RMSE} & \multirow{2}{*}{0.471} & \multirow{2}{*}{M\_RMSE} & \multirow{2}{*}{0.678} & \multirow{2}{*}{M\_RMSE} & \multirow{2}{*}{0.655} &  \\ \cline{2-2}
 & $n_{n}$ =256 &  &  &  &  &  &  &  \\ \cline{2-8}
 & $s_{b}$ = 128 & \multirow{2}{*}{Std\_RMSE} & \multirow{2}{*}{0.001} & \multirow{2}{*}{Std\_RMSE} & \multirow{2}{*}{0.04} & \multirow{2}{*}{Std\_RMSE} & \multirow{2}{*}{0.036} &  \\ \cline{2-2}
 & $r_{l}$ =0.0005 &  &  &  &  &  &  &  \\ \cline{1-8}
\multirow{4}{*}{\textbf{GC + HSEGA}} & $n_{f}$ =160 & \multirow{2}{*}{M\_RMSE} & \multirow{2}{*}{0.24} & \multirow{2}{*}{M\_RMSE} & \multirow{2}{*}{0.68} & \multirow{2}{*}{M\_RMSE} & \multirow{2}{*}{0.67} &  \\ \cline{2-2}
 & $n_{n}$ =192 &  &  &  &  &  &  &  \\ \cline{2-8}
 & $s_{b}$ =64 & \multirow{2}{*}{Std\_RMSE} & \multirow{2}{*}{0.02} & \multirow{2}{*}{Std\_RMSE} & \multirow{2}{*}{0.02} & \multirow{2}{*}{Std\_RMSE} & \multirow{2}{*}{0.02} &  \\ \cline{2-2}
 & $r_{l}$ =0.0013 &  &  &  &  &  &  &  \\ \cline{1-8}
\multicolumn{2}{|l|}{\begin{tabular}[c]{@{}l@{}}\textbf{T-test on results} 
\\with significance level of $\alpha$ = 5\%\end{tabular}} & \multicolumn{2}{l|}{\begin{tabular}[c]{@{}l@{}}$t$ = -59.840, $h$ = 1 \\ to reject the \\ equal mean hypothesis\end{tabular}} & \multicolumn{2}{l|}{\begin{tabular}[c]{@{}l@{}}$t$ = 0.745, $h$ = 0 \\ to accept the \\ equal mean hypothesis\end{tabular}} & \multicolumn{2}{l|}{\begin{tabular}[c]{@{}l@{}}$t$ = 1.816, $h$ = 0 \\ to accept the \\ equal mean hypothesis\end{tabular}} &  \\ \cline{1-8}
\end{tabular}
}%
\end{table*}

In tackling the Lipophilicity dataset, the results in Table \ref{tab:tab4 Results on Lipophilicity Data Set} show that the proposed approach is far better on the training dataset, and not worse than GC+BHO on validation and test datasets. As the M\_RMSE on the training set is less than that on the validation and test dataset, the proposed HESGA might have the over-fitting issue, which reduce its performance on validation and test datasets. Moreover, the Lipophilicity dataset has the biggest size among the three (more than 4,000 SMILE entries), so it introduces more complicated computational operations than the other datasets, which makes the execution very time-consuming.

\subsection{Experimental Results of HESGA to optimise MPNN Models}\label{section: 4.3}
As MPNN models is more time-consuming than GC, we only carried out experiments on FreeSolv dataset, and the detailed results are shown in Table \ref{tab:tab5 Results on FreeSolv Data Set 2}.

\begin{table*}[!t]
\centering
\caption{The Results on FreeSolv Dataset (2)}
\label{tab:tab5 Results on FreeSolv Data Set 2}
\resizebox{\textwidth}{!}{%
\begin{tabular}{|l|l|l|l|l|l|l|l|l}
\cline{1-8}
\textbf{FreeSolv} & \textbf{Hyperarameters} & \multicolumn{2}{l|}{\textbf{Training   Results}} & \multicolumn{2}{l|}{\textbf{Validation Results}} & \multicolumn{2}{l|}{\textbf{Test Results}} &  \\ \cline{1-8}
\multirow{4}{*}{\textbf{MPNN + BHO}} & T = 2 & \multirow{2}{*}{M\_RMSE} & \multirow{2}{*}{0.31} & \multirow{2}{*}{M\_RMSE} & \multirow{2}{*}{1.20} & \multirow{2}{*}{M\_RMSE} & \multirow{2}{*}{1.15} &  \\ \cline{2-2}
 & M = 5 &  &  &  &  &  &  &  \\ \cline{2-8}
 & $s_{b}$ = 16 & \multirow{2}{*}{Std\_RMSE} & \multirow{2}{*}{0.05} & \multirow{2}{*}{Std\_RMSE} & \multirow{2}{*}{0.02} & \multirow{2}{*}{Std\_RMSE} & \multirow{2}{*}{0.12} &  \\ \cline{2-2}
 & $r_{l}$ = 0.001 &  &  &  &  &  &  &  \\ \cline{1-8}
\multirow{4}{*}{\textbf{MPNN + HSEGA}} & T = 1 & \multirow{2}{*}{M\_RMSE} & \multirow{2}{*}{0.70} & \multirow{2}{*}{M\_RMSE} & \multirow{2}{*}{1.15} & \multirow{2}{*}{M\_RMSE} & \multirow{2}{*}{1.09} &  \\ \cline{2-2}
 & M = 10 &  &  &  &  &  &  &  \\ \cline{2-8}
 & $s_{b}$ = 8 & \multirow{2}{*}{Std\_RMSE} & \multirow{2}{*}{0.13} & \multirow{2}{*}{Std\_RMSE} & \multirow{2}{*}{0.15} & \multirow{2}{*}{Std\_RMSE} & \multirow{2}{*}{0.14} &  \\ \cline{2-2}
 & $r_{l}$ = 0.0012 &  &  &  &  &  &  &  \\ \cline{1-8}
\multicolumn{2}{|l|}{\begin{tabular}[c]{@{}l@{}}\textbf{T-test on results}\\
with significance level of $\alpha$ = 10\%\end{tabular}} & \multicolumn{2}{l|}{\begin{tabular}[c]{@{}l@{}}$t$ = 14.693, $h$   = 1 \\ to reject the \\ equal mean hypothesis\end{tabular}} & \multicolumn{2}{l|}{\begin{tabular}[c]{@{}l@{}}$t$ = -1.835, $h$   = 1 \\ to reject the \\ equal mean hypothesis\end{tabular}} & \multicolumn{2}{l|}{\begin{tabular}[c]{@{}l@{}}$t$ = -1.842, $h$  = 1 \\ to reject the \\ equal mean hypothesis\end{tabular}} &  \\ \cline{1-8}
\end{tabular}%
}
\end{table*}

As shown in Tab \ref{tab:tab5 Results on FreeSolv Data Set 2}, we carried out experiments on applying BHO and HESGA to optimise MPNN models. In terms of validation and test, the results show that there is no significant difference between the two sample groups with the significance level at 5\%; however, with the significance level of two tailed 5\% (i.e. 10\%), the equal mean hypothesis was rejected, which indicates that our algorithm is slightly better. Moreover, it is observed that on the training dataset the compared algorithm (MPNN + BHO) is far better than ours, which indicates that there may be a potential overfitting issue in that approach.

Overall, it seems that there are some cases of overfitting in the experiments (Tables \ref{tab:tab2 Results on ESOL Data Set (1)}$\sim$ \ref{tab:tab5 Results on FreeSolv Data Set 2}). The experimental results show that all RMSEs on the training datasets are less than those on validation and test. Particularly, the RMSE on the validation and test datasets is around two to four times than that on the training set in GC + BHO on the FreeSolv dataset and MPNN + BHO on the Lipophilicity dataset. As a result, overfitting might lead to poorer model performance  on validation/test datasets. For example, in Table \ref{tab:tab5 Results on FreeSolv Data Set 2}, the training loss of MPNN + BHO is just 50\% of that of MPNN + HESGA, but the loss of MPNN + BHO on the test and validation datasets are worse than that of MPNN + HESGA. 

%% file: Section5FurtherDiscussion.tex
\section{Further Discussions}
\label{section: 5 further discussion}
\subsection{The Distributions of RMSEs}\label{section:5.1}
\begin{figure*}[!t]
  \centering
  \begin{subfigure}[b]{0.4\linewidth}
    \includegraphics[width=\linewidth]{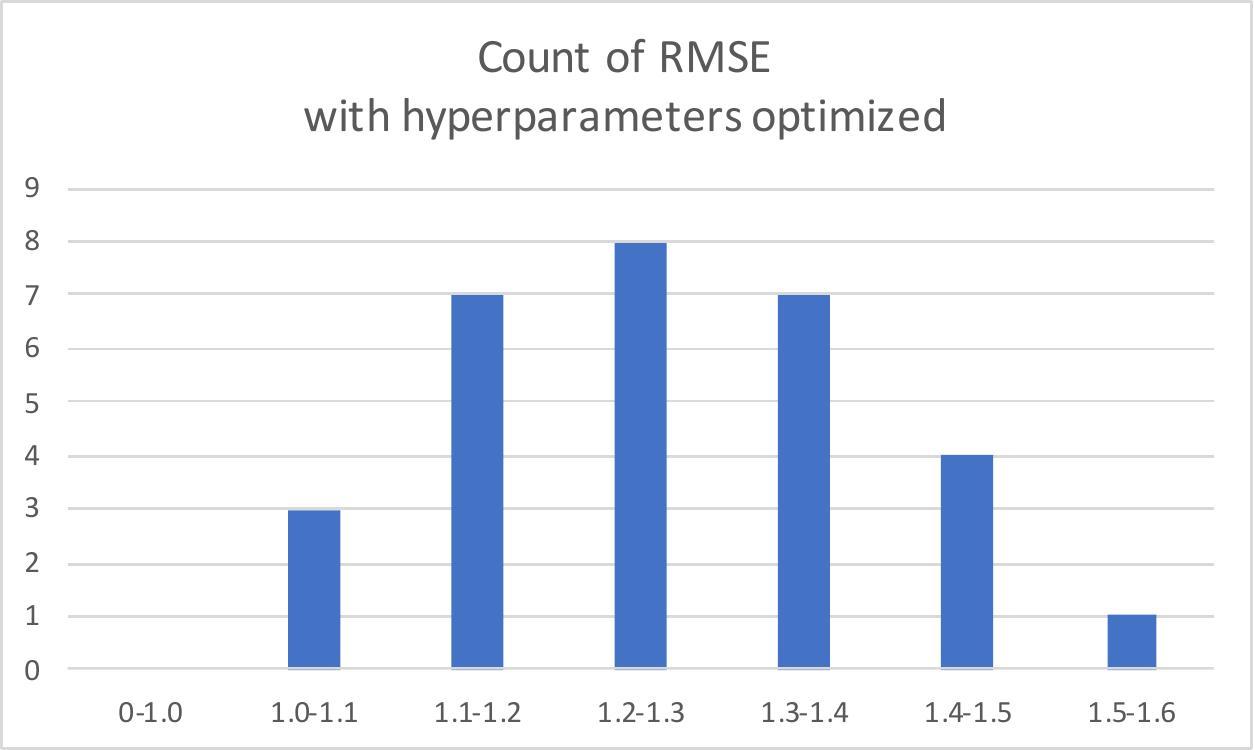}
    \caption{with the hyperparameter solution optimized}
  \end{subfigure}
  \begin{subfigure}[b]{0.4\linewidth}
    \includegraphics[width=\linewidth]{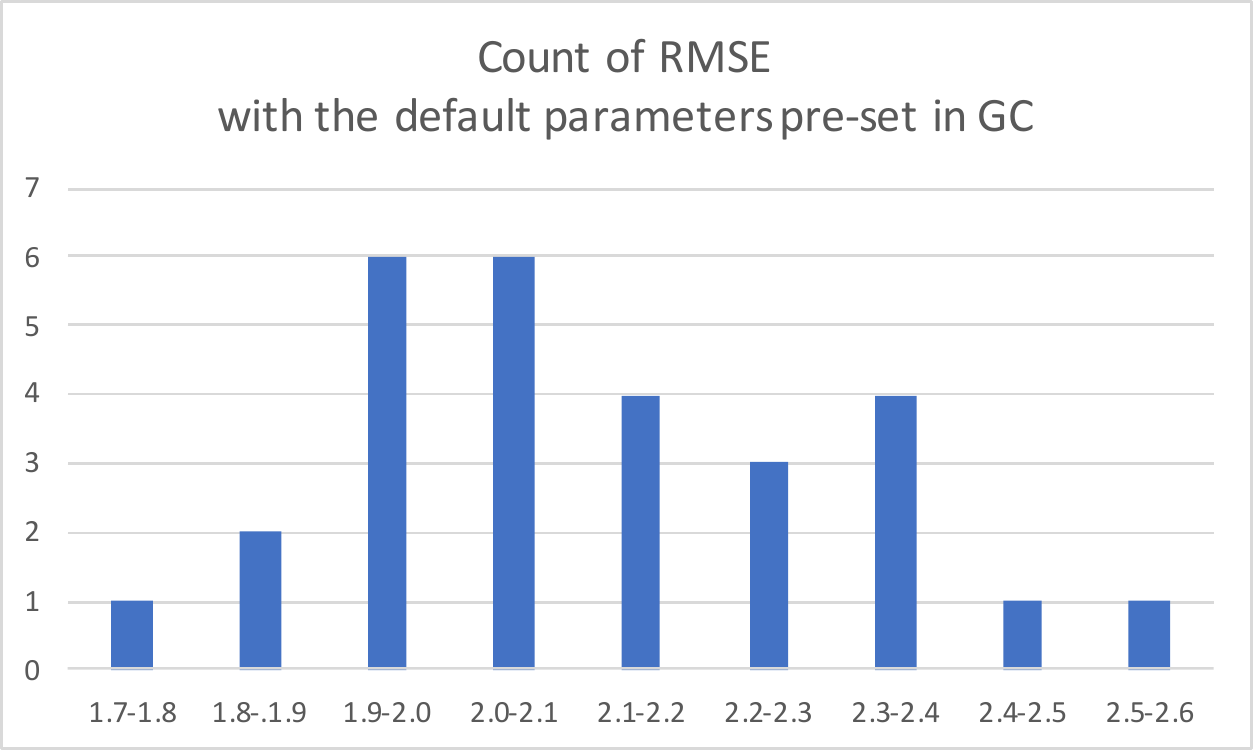}
    \caption{with the default parameters pre-set in GC}
  \end{subfigure}
  \caption{The Distribution of RMSE Results on the Validation Set by the Hyperparameters Optimized and the Default Hyperparameters Pre-set in GC}
  \label{fig:fig5}
\end{figure*}

Given the same set of hyperparameter values for a GNN model, the training results may be still different from time to time, even for the same split of the datasets, and this is mainly because in each training process, the weight vectors for a neural network are randomly initialised. As a result, GNN may produce variate RMSEs as a full evaluation function for GA, which increases uncertainty in evaluating all individuals in the GA population. Fig. \ref{fig:fig5} shows the distribution of RMSE results on the validation set under two given hyperparameter settings (one is the default parameters and the other is the hyperparameters optimised by HESGA).

As shown in Fig. \ref{fig:fig5}, the RMSE values are quite variable in 30 independent trials. One method to alleviate this negative effect is as follws: we performed experiments on using the average RMSE of several times (e.g. 3 times in a trial) of running GNN, however it will make the computational cost 3 times more expensive than before. And this is another reason that we need to develop a fast evaluation strategy for GA.

\subsection{Solution Resolution and Feasible Searching Space}
As presented in Section \ref{section:3.1}, with a higher resolution of hyperparameters being set up, we will have to deal with more feasible solution points. On one hand, lower resolution would alleviate the computational cost by reducing the number of feasible solutions, but it would be more likely to miss high quality solutions. On the other hand, a higher resolution of hyperparameter space will incur heavier computational overheads, but it would be more likely to identify a better set of hyperparameters compared with lower resolution. For comparison, we set up a series of experiments on the FreeSolv dataset with a varying resolutions of 8, 16, 32, 64 for encoding the batch size and the number of filters. In Table \ref{tab:figsolutionnumber}, we list the number of feasible solutions according to the different resolutions.

\begin{figure}[!t]
    \centering
    \includegraphics[width=3in]{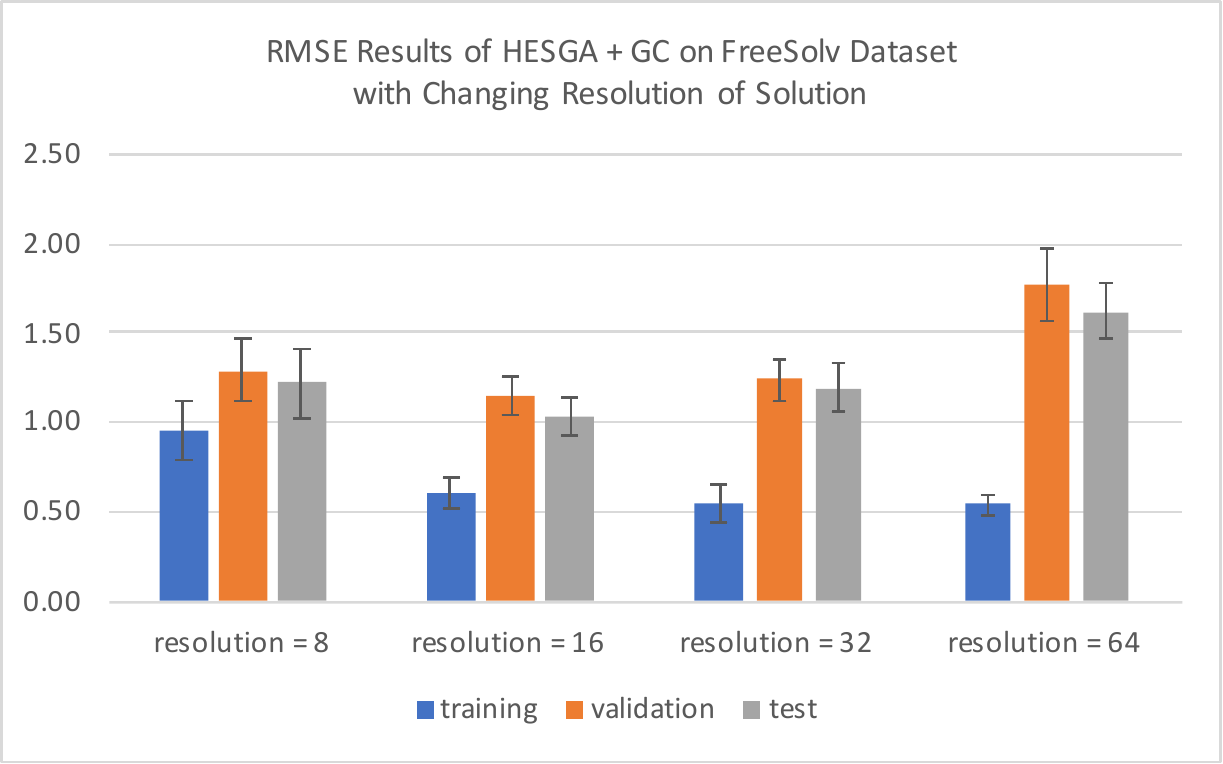}
    \caption{The RMSE Results of HESGA + GC on FreeSolv Dataset with Varing Resolutions of Hyperparameters}
    \label{fig:fig6RMSEResultsofHESGA+GConFreeSolvDatasetwithChangingResolutionofSolution}
\end{figure}

As shown in Fig. \ref{fig:fig6RMSEResultsofHESGA+GConFreeSolvDatasetwithChangingResolutionofSolution}, HESGA with lower resolution (bigger step increment) cannot find the optimal solutions found by those with higher resolution (smaller step increment). This is mainly because the grid generated in lower resolution is so coarse for this problem. On the other hand, the resolution of 16 will be acceptable for this problem as further higher resolution such as the resolution at 8 will not gain much improvement on the RMSE values. However, choosing an appropriate resolution may be a problem-specified issue; in the case of having abundant computational resources, we recommend to use as high resolution as possible to achieve better performance for GNN.

\begin{table}[]
\centering
\caption{Solution Number under Different  Resolution}
\label{tab:figsolutionnumber}
\resizebox{0.4\textwidth}{!}{%
\begin{tabular}{@{}lllll@{}}
\toprule
Resolution (step increment) & 8 & 16 & 32 & 64 \\ \midrule
Size of binary solution & 19 & 17 & 15 & 13 \\
Solution number & 524288 & 131072 & 32768 & 8192 \\ \bottomrule
\end{tabular}%
}
\end{table}

\subsection{Computational Cost} \label{section 5.3}
There are three processes that can affect the computational cost of the algorithms used in our experiments, \textit{1)} full GC evaluation, \textit{2)} fast evaluation of GC, and \textit{3)} HESGA. 

\subsubsection{Full Evaluation of GC}
Here we take the GC model as an example. As the GC model will first transfer a SMILE representation to a molecular fingerprint, we suppose that the fingerprints have a depth of $R$ and length of $L$, $N$ atoms were used in a molecular convolutional net \cite{duvenaud2015convolutional}, and $F$ features (filters) are used. In this case, in each layer the computational costs of feedforward and backpropagation process can be estimated by $O(\textit{RNFL} + \textit{RNF}^{2})$ \cite{duvenaud2015convolutional}. For simplicity, $O_{GC}$ stands for $O(\textit{RNFL} + \textit{RNF}^{2})$, which denotes the cost of GC model with one layer and one epoch of training.

\subsubsection{Fast Evaluation of GC}
As mentioned above, approx. 10\%$\sim$20\% of the maximum number of epochs were used to get the fast evaluation score, thus the cost will be 10\%$\sim$20\% of the full evaluation. As a result, for a number of $n_{e}$ epochs, the time cost will be $n_e \times O_{GC}$ approximately. Suppose the fast evaluation will use a percentage of $p_{f}$ of the total epochs, and in this case the cost of fast evaluation will be $ p_{f} \times n_{e} \times O_{GC}$. 

\subsubsection{Total Cost of HESGA}
Suppose we have a GC with one convolution layer, a population of solutions with size $n_{pop}$, and each individual is trained by $n_{e}$ epoch at most, the proportion of elite group is $r_{e}$, and the maximum generation is $maxgen$. The detailed cost of HESGA is listed as follows based on Algorithm \ref{alg1}:

\textbf{Lines 1$\sim$3:} for full evaluations on the whole population, it will cost $n_{pop} \times n_{e} \times O_{GC}$ approximately. 

\textbf{Lines 4$\sim$9:} in one generation, for fast evaluation on the whole new offspring, it will cost $n_{pop} \times p_{f} \times n_{e} \times O_{GC}$; for the full evaluation on candidates, it will cost $n_{pop} \times r_{c} \times n_{e} \times O_{GC}$, thus totally the total cost will be $(p_{f}+ r_{c}) \times n_{pop} \times n_{e} \times O_{GC}$. Therefore, for $maxgen$ generations, it will cost $(p_{f}+ r_{c})  \times n_{pop} \times n_{e} \times O_{GC} \times maxgen$. It should be noted here that the cost of sorting and counting operations can be ignored compared with $O_{GC}$.

As a result, HESGA will cost approximately $[(p_{f}+ r_{c}) \times maxgen+1] \times n_{pop} \times n_{e} \times O_{GC}$. Take an example as follows: suppose we have $n_{pop} = 10$, $maxgen = 10$, $p_{f} = 0.1$, $r_{c} = 0.1$, $n_{e}= 100$. In this case, thus running HESGA once will equal to running 3,000 times of single $O_{GC}$ in terms of the computational cost.

\subsection{The Scalability of HESGA} \label{section 5.4}
We argue that HESGA possesses scalability for different problems and datasets. For example, for a large dataset, the fast evaluation approach can be replaced by any reasonable approaches, such as partial training by using sampled data points randomly from the whole dataset, as in \cite{real2019regularized}. When historical datasets are available, a fast surrogate model could be built and trained with the datasets, to approximate the results obtained from the complete training. No matter what types of fast evaluation approaches are used, the HESGA will always be a good mechanism to combine both fast and full evaluation to achieve a trade-off between solution quality and computational cost. 

% We argue that HESGA is scalable to larger search spaces. The fast evaluation approach can be replaced by any reasonable approaches, e.g. partial training by using sampled data points randomly from the whole dataset, as in \cite{real2019regularized}. Another example is building a fast surrogate model by training the model on historical datasets, and then use this trained model to replace the fast evaluation proposed in this research. No matter what types of fast evaluation approaches are used, the HESGA will always be a good mechanism to combine both fast and full evaluation to achieve a trade-off between solution quality and computational cost.

%% file: Section6Conclusion.tex
\section{Conclusion and Future Work}
\label{section: 6 conclusion}
In this research, we proposed HESGA, a novel GA equipped with a hierarchical evaluation strategy and full and fast evaluation methods, is proposed to address expensive the HPO problems for GNNs. Experiments are carried out on three representative datasets in material property prediction problems: ESOL, FreeSolv, and Lipophilicity datasets; by applying HESGA to optimise the hyperparameters of GC and MPNN models, two types of commonly used graph deep neural networks in material design and discovery. Results show that HESGA can outperform BHO when optimising GC models, meanwhile it achieves comparable performance to Bayesian approaches to optimising MPNN models. In Section 5, we also analysed the uncertainty and distributions of RMSE results, the learning performance in terms of the resolution of the hypereparameter search space, the computational cost, and the scalability of HESGA. 

In the future, we would like to investigate the following two aspects:

    \paragraph{\textbf{Dealing with the over-fitting issue in the experiments}}
    This is an issue observed in both the Bayesian approaches and our HESGA. In our experiments, the number of epochs ($n_{e}$) is not specified as one hyperparameter, which might be one reason for overfitting. For an example, overtraining might cause HPO biases to a perfect fitted model on the training dataset but this model may perform poorly in validation and test datasets. Therefore, from our perspective, we would like to investigate how we can incorporate more hyperparameters in the search space, or to monitor the overfitting and introduce the penalty item in the evaluation functions. 

    \paragraph{\textbf{Bi-objective Optimization}} 
    As mentioned in Section \ref{section 5.3}, the hyperparameters such as the number of epochs ($n_{e}$) and the number of filters ($n_{f}$)  are selected to be optimized, and this will affect the computational cost of HESGA. Suppose the RMSE might be improved while the cost would be increased at the same time when we increase $n_{e}$ and $n_{f}$, and in this case a balance between the performance and cost needs to be considered. In our future work, we will consider dealing with this balance issue as a bi-objective optimization problem, and a Pareto-optimal front (PF) \cite{deb2002fast} is expected to offer more options of GNN models considering the trade-off between performance and cost. 

%% file: Section7Acknowledge.tex
\section{Acknowledgement}
This research is supported by the Engineering and Physical Sciences Research Council (EPSRC) funded Project on New Industrial Systems: Manufacturing Immortality (EP/R020957/1). The authors are also grateful to the Manufacturing Immortality consortium.